\documentclass{article}

\PassOptionsToPackage{numbers, compress, sort}{natbib}

\usepackage[preprint]{neurips_2026}

\usepackage[utf8]{inputenc} 
\usepackage[T1]{fontenc}    
\usepackage{hyperref}       
\usepackage{url}            
\usepackage{booktabs}       
\usepackage{amsfonts}       
\usepackage{nicefrac}       
\usepackage{microtype}      
\usepackage{xcolor}         
\usepackage{subcaption}
\usepackage[dvipsnames]{xcolor}

\usepackage{amsmath,amssymb,amsthm,mathtools,bm}
\usepackage{multirow}
\usepackage{graphicx}
\usepackage[table]{xcolor}
\usepackage{caption}
\usepackage{array}
\usepackage{wrapfig}
\makeatletter
\renewcommand\thesubtable{(\alph{subtable})}
\makeatother

\title{Learning to Solve Generative ODEs \\
Beyond the Linear Span}

\author{%
\textbf{Sihyeon Kim}$^{1}$ \quad
\textbf{Seunghun Lee}$^{2}$ \quad
\textbf{Vikas Singh}$^{3,\S}$ \quad
\textbf{Hyunwoo J. Kim}$^{2,\S}$\\
$^{1}$Korea University \quad
$^{2}$KAIST \quad
$^{3}$University of Wisconsin--Madison\\
{\small
\texttt{sh\_bs15@korea.ac.kr} \quad
\texttt{\{llsshh319,hyunwoojkim\}@kaist.ac.kr} \quad
\texttt{vsingh@biostat.wisc.edu}
}
}

\begin{document}

\maketitle

\begingroup
\renewcommand{\thefootnote}{\S}
\footnotetext{Equal advising.}
\endgroup

\newcommand{\tableCellHeight}{1}
\newcommand{\tabstyle}[1]{
  \setlength{\tabcolsep}{#1}
  \renewcommand{\arraystretch}{\tableCellHeight}
  \centering
  \small
}

\definecolor{tabhighlight}{HTML}{e5e5e5}
\definecolor{mod}{rgb}{0,0,0} 
\definecolor{mod2}{rgb}{1,0.1,0.1} 
\definecolor{sotacolor}{rgb}{0.88, 0.93, 1.0}
\definecolor{paramcolor}{rgb}{0.9, 0.2, 0.5}  
\definecolor{commentcolor}{rgb}{0.0, 0.0, 1.0} 
\definecolor{color_blue}{HTML}{2E86AB}  
\definecolor{color_red}{HTML}{A23B72}  

\newtheorem{theorem}{Theorem}[section]
\newtheorem{proposition}[theorem]{Proposition}
\newtheorem{remark}[theorem]{Remark}
\newtheorem{assumption}[theorem]{Assumption}
\newtheorem{corollary}[theorem]{Corollary}

\newcommand{\Proj}{\Pi}
\newcommand{\Risk}{\mathcal{R}}
\newcommand{\Slin}{\mathcal{S}_{\mathrm{lin}}}
\newcommand{\Sop}{\mathcal{S}_{\mathrm{op}}}
\newcommand{\eg}{\textit{e.g.}}
\newcommand{\ie}{\textit{i.e.}}

\vspace{-5pt}
\begin{abstract}
Diffusion and flow generative models sample by integrating a learned ODE, but high quality still requires many sequential model evaluations. 
Solver learning reduces this cost by adapting scalar coefficients, timesteps, or both, while keeping the backbone model fixed. 
In this work, we identify a structural bottleneck in this update family: each step remains span-limited.
Since the scalar-coefficient update lies in the span of buffered velocity evaluations, it can fit only the in-span component while leaving any out-of-span residual unreachable by scalar recombination alone.
We propose SpanLift, a lightweight neural solver that augments scalar-coefficient updates with a spatial residual operator.
SpanLift keeps a fixed base solver as an in-span prior and learns a spatial residual operator over the state and velocity buffer. 
The operator is trained by endpoint teacher matching, preserves the pretrained backbone, and adds no model NFEs. 
Empirically, the learned correction transfers across base solvers and is predominantly out-of-span. 
Across pixel-space diffusion, latent flow matching, and precipitation nowcasting, SpanLift achieves state-of-the-art few-step sampling. 
With only 3 NFE, it improves CIFAR-10 FID from 8.16 to 5.69 and ImageNet FID from 17.37 to 11.83.
\end{abstract}
\section{Introduction}
\label{sec:intro}
\vspace{-5pt}

Diffusion and flow generative models have become a powerful backbone for high-quality synthesis, leading to extensive use in content generation~\cite{DDPM,EDM,flux,xie2025sana,wan2025} as well as scientific applications, from protein design~\cite{watson2023rfdiffusion,bose2024foldflow} to weather forecasting~\cite{gao2023prediff}.
Despite the high quality of outputs from these models, inference remains costly because generating each example requires many evaluations of the learned score/velocity model along the generative ODE.
To close this gap, model distillation methods~\cite{salimans2022pd,song2023cm,song2024ict} train a student model to approximate multi-step sampling with fewer evaluations.
However, they require substantial training cost and may alter the behavior of the original model.
Another alternative is to keep the model frozen and improve the numerical solver. 
Analytic multistep solvers~\cite{dpmpp,PNDM,deis,UniPC} improve sampling efficiency by combining current and buffered model outputs without extra model evaluations, which can be viewed as quadratures over the learned velocity field~\cite{kharazmi2021hp,palpinns}.
Here, analytically derived coefficients rely on small-step assumptions.
In the low NFE regime, this premise breaks down, since large steps render low-degree local approximations unreliable~\cite{s4s}, and fixed coefficients cannot adapt to the nonstationary nature of the sampling dynamics~\cite{dyweight}.

The limitations of analytic solvers have motivated ``solver learning'', which learns the sampling rule for integrating the generative ODE with only a few model evaluations.
These methods usually follow a teacher-student setup, where a high-accuracy teacher sampler provides reference outputs and a lower NFE student solver is trained to reproduce them, either through trajectory-level or rollout-level supervision.
Within this framework, existing work mainly adapts two solver choices.
One cluster of works learns step-dependent coefficients for combining current and buffered model outputs, replacing analytically derived quadrature weights~\cite{bns,s4s,dyweight}.
Another set of results learn where to query the model along the trajectory, by optimizing timestep schedules or intermediate evaluation points~\cite{LD3,AMED}.
Such work~\cite{s4s,dyweight,DSS} exists to learn both.
This lets the solver adapt to large step sampling, where analytic quadrature becomes unreliable.

\vspace{-2pt}
\textbf{Observation.} This raises a natural question: under an extreme NFE budget, can better scalar combinations or quadrature points alone close the gap to the teacher?
Despite differing in what they learn, existing methods still execute each step by scalar-recombining the available velocity evaluations, confining the update to the velocity-buffer span.
We observe through a controlled ODE analysis (Figure~\ref{fig:toy_analysis}) that such local confinement leaves a systematic gap.
Decomposing the update mismatch into an in-span component and an orthogonal out-of-span residual, we find that scalar-coefficient learning can reduce the former by moving toward the best in-span projection, but cannot reduce the latter with \emph{any} scalar recombination.
Few-step sampling is therefore not only a matter of optimizing scalar coefficients or evaluation points.
Once the velocity history at a step is fixed, further reducing the out-of-span residual calls for enlarging the class of admissible solver updates beyond the velocity-buffer span.

\vspace{-2pt}
\textbf{This paper.} We propose \emph{SpanLift}, a solver framework that augments scalar-coefficient updates with a lightweight spatial operator.
The base solver preserves the reliability of existing multistep samplers, while the learned operator supplies residual corrections beyond the history span, using neighboring states and model outputs to capture spatially coupled directions suggested by the local structure of generative ODEs. 
SpanLift uses endpoint teacher matching with the generative model frozen, adding {\em no model evaluations beyond lightweight residual operator overhead}.
Across image generation with pixel-space diffusion and latent flow-matching models, as well as precipitation nowcasting, our solver achieves state-of-the-art few-step results with large margins, with larger gains under more extreme NFE budgets.
In addition, SpanLift consistently improves diverse base solvers, from Runge--Kutta to scalar-coefficient multistep solvers, indicating the learned operator captures correction structure not tied to a particular update rule.
These results show that treating the sampler itself as a lightweight neural solver provides a general and efficient route to stronger few-step sampling.

\begin{figure}[!b]
    \centering
\vspace{-17pt}

    \begin{subfigure}{0.48\linewidth}
        \centering
        \begin{tabular}{@{}cc@{}}
            \small DyWeight & \small SpanLift \\
            \includegraphics[width=0.48\linewidth]{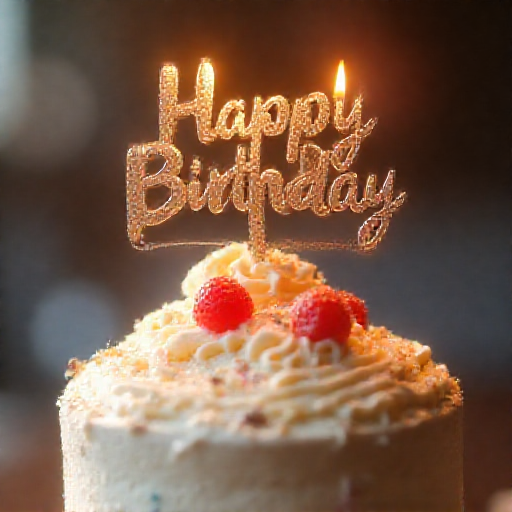} &
            \includegraphics[width=0.48\linewidth]{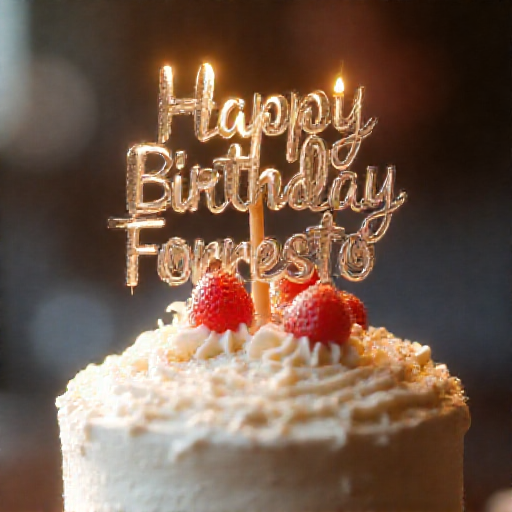}
        \end{tabular}
        \caption{A birthday cake with a topper that reads \\ \hspace*{1.3em}\textbf{"Happy Birthday Forresto."}}
    \end{subfigure}
    \hfill
    \begin{subfigure}{0.48\linewidth}
        \centering
        \begin{tabular}{@{}cc@{}}
            \small DyWeight & \small SpanLift \\
            \includegraphics[width=0.48\linewidth]{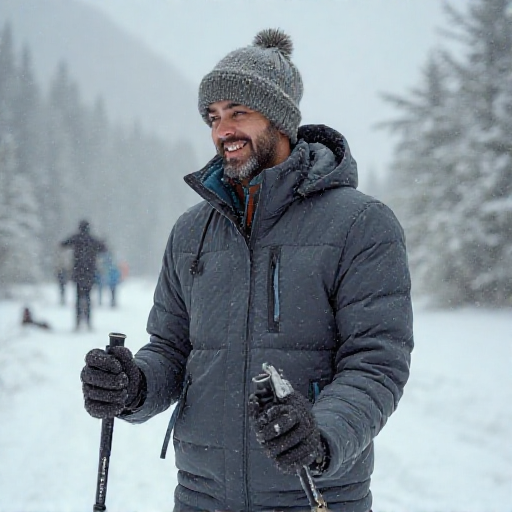} &
            \includegraphics[width=0.48\linewidth]{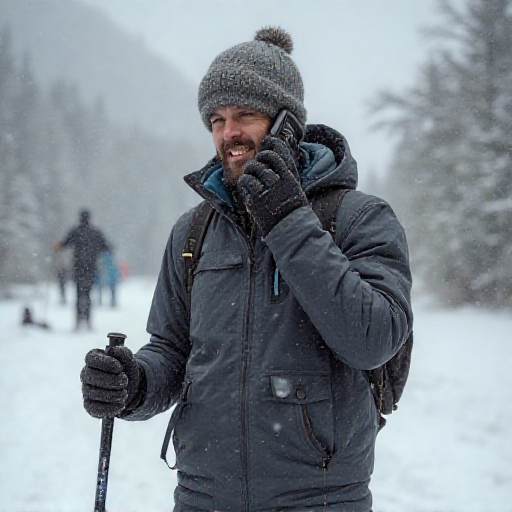}
        \end{tabular}
        \caption{A man \textbf{talking on a phone} while holding ski poles.\\}
    \end{subfigure}

    \vspace{-3pt}
    \caption{Qualitative results generated with FLUX.1-dev~\cite{flux} ((a) NFE=5, (b) NFE=9).}
    \label{fig:flux_nfe9_qual}
    \vspace{-12pt}
\end{figure}

\vspace{-5pt}
\begin{figure}[!b]
    \centering
    \begin{subfigure}{0.30\linewidth}\phantom{Xg}\end{subfigure}\hfill
    \begin{subfigure}{0.21\linewidth}\centering\caption*{iPNDM}\end{subfigure}\hfill
    \begin{subfigure}{0.21\linewidth}\centering\caption*{DyWeight}\end{subfigure}\hfill
    \begin{subfigure}{0.21\linewidth}\centering\caption*{SpanLift}\end{subfigure}
    \vspace{-0pt}
    \begin{subfigure}[c]{0.30\linewidth}
        \raggedright\footnotesize\itshape
        ``Create a realistic image of a scenery of a Minotaur that is dressed as a barkeeper, standing behind a counter in a busy Tavern. The Minotaur is polishing a small glass. The ambience should be in Dungeons and Dragons Style''
    \end{subfigure}\hfill
    \begin{subfigure}[c]{0.21\linewidth}
        \centering\includegraphics[width=\linewidth]{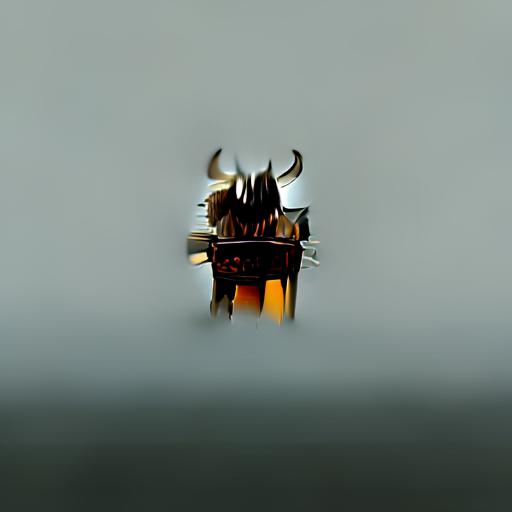}
    \end{subfigure}\hfill
    \begin{subfigure}[c]{0.21\linewidth}
        \centering\includegraphics[width=\linewidth]{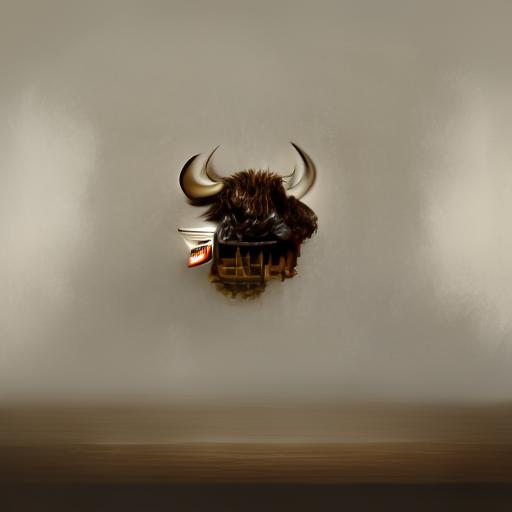}
    \end{subfigure}\hfill
    \begin{subfigure}[c]{0.21\linewidth}
        \centering\includegraphics[width=\linewidth]{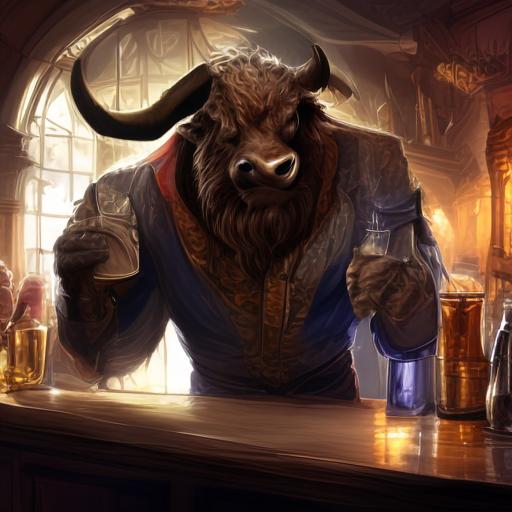}
    \end{subfigure}

    \vspace{1pt}
    \begin{subfigure}[c]{0.30\linewidth}
        \raggedright\footnotesize\itshape
        ``a picture of a room with ginormous floor to ceiling windows. There is a huge oversized couch. green lush plants are everywhere and hang from the ceiling. Plants are also on shelves on the walls. There is a cute coffee table. Bohemian style. \ldots''
    \end{subfigure}\hfill
    \begin{subfigure}[c]{0.21\linewidth}
        \centering\includegraphics[width=\linewidth]{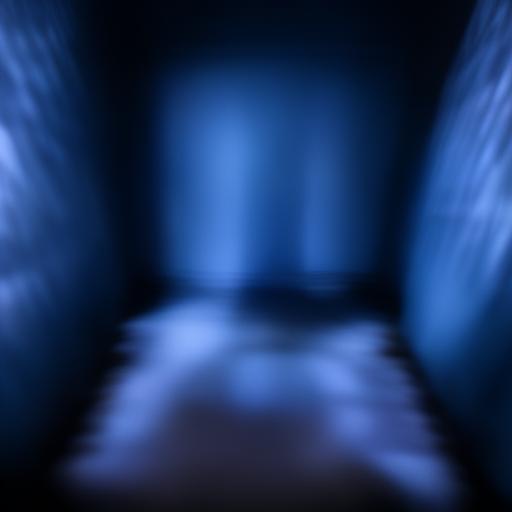}
    \end{subfigure}\hfill
    \begin{subfigure}[c]{0.21\linewidth}
        \centering\includegraphics[width=\linewidth]{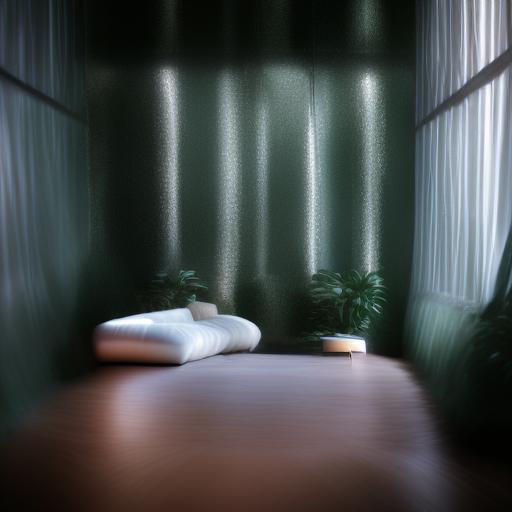}
    \end{subfigure}\hfill
    \begin{subfigure}[c]{0.21\linewidth}
        \centering\includegraphics[width=\linewidth]{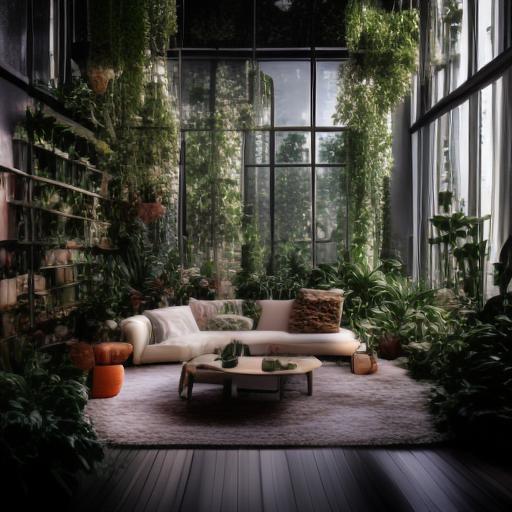}
    \end{subfigure}

    \caption{Qualitative results generated with SANA-0.6B~\cite{xie2025sana} (NFE=5).}
    \label{fig:sana_nfe5_qual}
\end{figure}
\section{Preliminaries}
\vspace{-5pt}
We consider deterministic sampling from a pretrained diffusion or flow-matching model in continuous time, starting from a noise sample at $t=0$ and evolving toward the clean sample at $t=1$.
Sampling follows the ODE $\dot{x}(t) = u_t(x(t))$, $t \in [0,1]$, with initial condition $x(0) = x_0 \sim p_0$, where
\begin{equation}
    u_t(x) \;=\; \mu_t\, x \;+\; \gamma_t\, f_\theta(x,t)
    \label{eq:velocity_field}
\end{equation}
is the time-dependent velocity field induced by the frozen pretrained network $f_\theta$.
The scalar functions $\mu_t, \gamma_t$ are determined by the noise schedule, $\eg$, for flow-matching models, $\mu_t = 0$ and $\gamma_t = 1$.
The sampling objective is to numerically integrate this ODE and obtain the clean sample $x(1)$.
\vspace{-2pt}
For any two times $t_i < t_{i+1}$, the exact solution satisfies
$x(t_{i+1}) = x(t_i) + \int_{t_i}^{t_{i+1}} u_t(x(t))\,dt$.
In practice, we introduce a time discretization $0 = t_0 < t_1 < \cdots < t_N = 1$ and compute an approximate trajectory $x_i \approx x(t_i)$ by replacing the integral with a computable approximation.
A generic solver step takes the increment form
\begin{equation}
    x_{i+1} = x_i + h_i\,\Delta_i,
    \qquad
    h_i := t_{i+1} - t_i,
    \label{eq:solver_step}
\end{equation}
where $\Delta_i$ denotes the solver update at step~$i$.
Standard solvers include Euler, Heun, and higher-order Runge--Kutta (RK) methods.
Higher-order RK variants reduce local discretization error via additional evaluations, but in the few-step regime, we find that these quickly consume the limited NFE budget.

\vspace{-4pt}
\subsection{Multistep solvers for few-step sampling}
\vspace{-5pt}

Rather than spending additional network evaluations at each step, modern few-step samplers~\cite{dpmsolver,dpmpp,PNDM,deis,UniPC,tan2026stork} reuse a buffer of previously computed model outputs to construct each update.
In their multistep form, the update direction is a linear combination of buffered velocity evaluations.

Let $u_i := u_{t_i}(x_i)$ denote the model velocity at the current numerical state.
A $K$-step multistep update is written as
\begin{equation}
    \Delta_i \;=\; \sum_{j=0}^{K-1} w_{i,j}^{\rm ana}\, u_{i-j},
    \label{eq:multistep_update}
\end{equation}
where $\{u_{i-j}\}_{j=0}^{K-1}$ are the buffered velocity fields from the current and past $K{-}1$ steps.
For example, iPNDM~\cite{deis} can be viewed as an extrapolatory polynomial quadrature rule over the velocity buffer, where its coefficients reduce to the classical Adams--Bashforth weights under uniform discretization,
forming the increment by extrapolating from three historical velocity fields.
The multistep variant of DPM-Solver++~\cite{dpmpp} also takes the form~\eqref{eq:multistep_update}, with step-dependent coefficients derived from the semi-linear structure of the diffusion ODE in logSNR parameterization.
In these handcrafted solvers, $w_{i,j}$ are derived \textit{analytically} from the chosen solver family and time discretization. 

\vspace{-5pt}
\subsection{Learnable multistep solvers for few-step sampling}
\vspace{-5pt}
Recently, learnable approaches~\cite{shaul2024bespoke, bns,learningfast,s4s,D-ODE,dyweight,LD3,timesteptuner,DMN,ays,AMED,EPD,park2026dualsolver,min2026bezierflow} have been proposed in place of fully handcrafted solvers.
These methods aim to improve few-step sampling by learning better solver design choices, either by learning the \textbf{better combination} of buffered velocities, or by learning \textbf{better quadrature points}, $\ie$, a better timestep schedule, or both.

\vspace{-5pt}
\paragraph{Learning solver parameters.}
These methods commonly use a teacher-student framework, where a high-accuracy teacher solver produces reference samples and a lower-NFE student solver is trained to reproduce them.
The methods differ in \emph{where} the teacher supervision is applied.
One line of work~\cite{AMED,EPD,DSS,DLMS,shaul2024bespoke} provides trajectory-level supervision, training the student to match the teacher at each coarse-grid time $t_i$.
However, recent results~\cite{LD3,s4s,dyweight} show that it can be overly restrictive in the low-NFE regime.
With large step sizes, the student may achieve a better final output by {\em deviating} from the teacher trajectory at intermediate times.
 
Thus, an alternative approach~\cite{bns,LD3,s4s,dyweight,min2026bezierflow} supervises only the \emph{final rollout endpoint}.
Given a student solver $S$ induces a rollout map $\Phi_S : x_0 \mapsto \hat{x}_1^S$ by recursively applying its update rule over $N$ steps, and a teacher rollout map $\Phi^\star$ with a finer discretization, the student is trained to match the teacher's final output:
\begin{equation}
 S_N^\dagger = \arg\hspace{-4pt}\min_{S \in \mathcal{S}_N} \mathcal{J}(S), \qquad
\mathcal{J}(S) :=
\mathbb{E}_{x_0 \sim p_0}
\bigl[
\ell\bigl(\Phi_S(x_0),\, \Phi^\star(x_0)\bigr)
\bigr],
\label{eq:global_objective}
\end{equation}
where $\ell$ is the distance between the student and teacher outputs.

\textcolor{mod}{
Note that endpoint supervision is applied only after the recursive rollout, thus does not define a unique target update at each coarse step.
Different per-step update sequences can reach the same endpoint without corresponding to teacher-trajectory velocities.
}

\vspace{-5pt}
\paragraph{What is learned?}
Existing methods optimize one or both of two solver components.
The first is the combination of buffered velocities.
Methods such as DLMS~\cite{DLMS}, S4S~\cite{s4s}, and DyWeight~\cite{dyweight} learn step-dependent scalar coefficients $W := [w_{i,j}] \in \mathbb{R}^{N \times K}$,
\begin{equation}
\Delta_i(W) = \sum_{j=0}^{K-1} w_{i,j}^{\rm learn}\,u_{i-j}.
\label{eq:learned_multistep_update}
\end{equation}
The second is the timestep schedule, \ie, the quadrature points at which the solver is executed, as in GITS~\cite{gits}, AYS~\cite{ays}, and LD3~\cite{LD3}.
More recent methods optimize both jointly, $\eg$, S4S-Alt~\cite{s4s} alternates between coefficient and schedule learning, while DyWeight~\cite{dyweight} parameterizes both the velocity combination and effective step size through a shared weight table.
Despite these differences, all proposals above still execute each step as a scalar-weighted combination of the velocity evaluations {\em available at that step}. 
\vspace{-4pt}
\section{Beyond scalar-coefficient updates}
\vspace{-5pt}

\subsection{Linear span limitation}
\label{sec:limitation}
\begin{figure}
    \centering
    \includegraphics[width=1\linewidth]{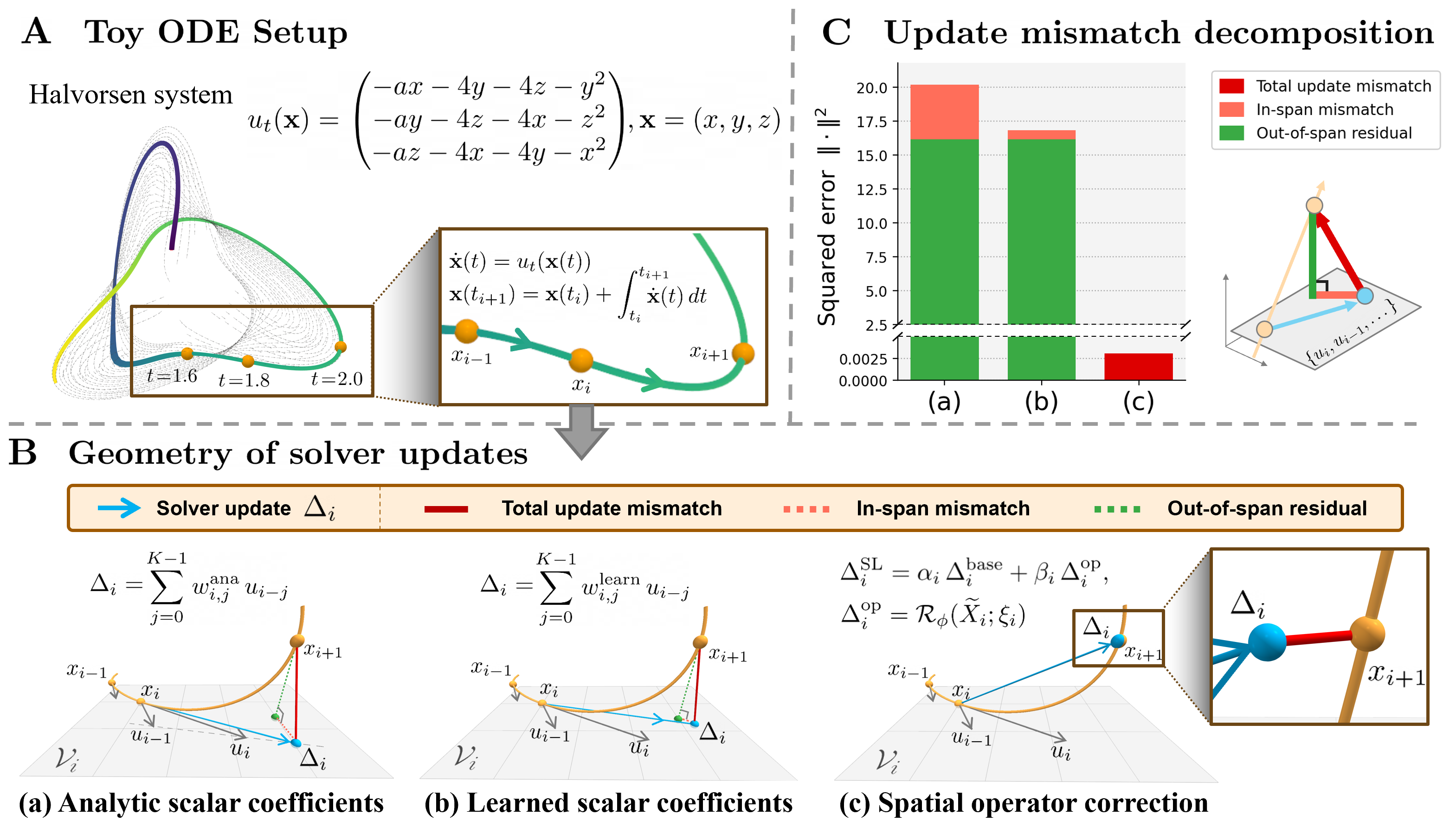}
    \vspace{-14pt}
    \caption{
    \textbf{3D ODE analysis of the span limitation.}
    \textbf{(A)} We use the Halvorsen system as a controlled nonlinear ODE and study a one-step update from \(t_i=1.8\) to \(t_{i+1}=2.0\).
    \textbf{(B)} With \(K=2\), the buffered velocities \(\{u_i,u_{i-1}\}\) induce the local span \(\mathcal V_i\).
    The scalar-coefficient updates in (a) Adams--Bashforth~\cite{hairer1993solving} and (b) DyWeight~\cite{dyweight} remain inside \(\mathcal V_i\), whereas (c) SpanLift can move outside the span.
    \textbf{(C)} The squared update mismatch decomposes as in~\eqref{eq:pythagorean_decomposition}.
    Compared with (a) and (b), (c) moves closer to the best in-span projection \(\Pi_{\mathcal V_i}\Delta_i^\star\), reducing the in-span mismatch.
    \textcolor{mod}{
    For a fixed state and velocity buffer, the out-of-span residual $\|r_i^\perp\|^2$ is a lower bound on the local error of any scalar-coefficient update.
    By moving beyond $\mathcal{V}_i$, SpanLift can reduce this residual along directions unavailable to any scalar combination.}
    }
    \vspace{-15pt}
    \label{fig:toy_analysis}
\end{figure}

Scalar-coefficient solvers can improve the choice of weights over the velocity history, but their updates remain confined to the span induced by that history.
We analyze this restriction by decomposing a reference update into its in-span component and the remaining out-of-span residual.
\vspace{-8pt}
\textcolor{mod}{
\paragraph{History-induced local span.}
At solver step $i$, both the handcrafted update~\eqref{eq:multistep_update} and the learned update~\eqref{eq:learned_multistep_update} take the form $\Delta_i = \sum_{j=0}^{K-1} w_{i,j}\, u_{i-j}$.
We collect these velocity vectors into the local basis set $\mathcal{B}_i := \{u_i,\, u_{i-1},\, \ldots,\, u_{i-K+1}\}$ and define its induced local linear span $\mathcal{V}_i := \operatorname{span}(\mathcal{B}_i)$.
Every scalar-coefficient update satisfies $\Delta_i \in \mathcal{V}_i$, regardless of whether the coefficients are fixed analytically or learned from data. 
We write $\Pi_{\mathcal{V}_i}$ for the orthogonal projection onto $\mathcal{V}_i$.}
\vspace{-5pt}
\textcolor{mod}{\paragraph{Local span lower bound.}
Since endpoint matching neither defines unique per-step targets nor pins the student to the teacher trajectory, we analyze the exact one-step continuation from the student's own state $x_i$.
Given $x_i$ and step size $h_i$, define the one-step teacher target
\begin{equation}
\Delta_i^{\star}(x_i)
:=
\frac{\Phi^{\star}_{t_i \to t_{i+1}}(x_i)-x_i}{h_i},
\label{eq:one_step_teacher_update}
\end{equation}
where $\Phi^{\star}_{t_i \to t_{i+1}}$ is the one-step restriction of the teacher map $\Phi^{\star}$ in~\eqref{eq:global_objective}.
This target is used only as an analytical reference, not as training supervision; the solver is still optimized end-to-end through~\eqref{eq:global_objective}.
Let $r_i^\perp := (I-\Pi_{\mathcal V_i})\Delta_i^\star$ denote the component of the target outside $\mathcal{V}_i$.
Since $r_i^\perp \perp \mathcal{V}_i$ and every scalar-coefficient update satisfies $\Delta_i\in\mathcal{V}_i$, the Pythagorean identity gives
\begin{equation}
\underbrace{\|\Delta_i^{\star} - \Delta_i\|^2}_{\text{update mismatch}}
\;=\;
\underbrace{\|\Pi_{\mathcal{V}_i} \Delta_i^{\star} - \Delta_i\|^2}_{\text{in-span mismatch}}
\;+\;
\underbrace{\|r_i^{\perp}\|^2}_{\text{out-of-span residual}}.
\label{eq:pythagorean_decomposition}
\end{equation}
The in-span mismatch vanishes at $\Delta_i=\Pi_{\mathcal{V}_i}\Delta_i^\star$, which is attainable since $\Pi_{\mathcal{V}_i}\Delta_i^\star\in\mathcal{V}_i$.
Therefore,
\begin{equation}
\min_{\Delta_i\in\mathcal{V}_i}
\|\Delta_i^{\star}-\Delta_i\|^2
=
\|r_i^{\perp}\|^2 .
\label{eq:local_span_lower_bound}
\end{equation}
See Figure~\ref{fig:toy_analysis} for a geometric illustration.
The first term is the in-span mismatch, which learned-coefficient methods~\cite{DLMS,s4s,dyweight,DSS} can reduce within $\mathcal{V}_i$.
The second term, $\|r_i^\perp\|^2$, is inaccessible to any scalar combination of the fixed velocity buffer $\mathcal{B}_i$, and is therefore a local error floor for the scalar-coefficient update class when approximating $\Delta_i^\star(x_i)$.
This local statement does not preclude schedule or endpoint learning from changing future states and buffers.
Since $\dim(\mathcal{V}_i)\le K$ while $\Delta_i^\star$ lies in the full update space, this floor can be substantial in few-step sampling, motivating learning updates beyond $\mathcal{V}_i$.}

\subsection{From scalar coefficients to learned operators}
\label{sec:operator_hierarchy}
\vspace{-6pt}
Next, we enlarge the scalar-coefficient update family by viewing solver steps as operators over the velocity buffer.
This operator view includes scalar combinations, pointwise maps, and spatial operators as increasingly expressive update classes, providing a compact design space beyond \(\mathcal V_i\) and motivating spatial kernels as a lightweight inductive bias for residuals with cross-position structure.

\vspace{-6pt}
\paragraph{Operator view of solver updates.}
We view each solver step as an operator over the available velocity buffer $X_i := (u_i,u_{i-1},\ldots,u_{i-K+1}).$
The scalar-coefficient update is recovered as
\begin{equation}
    \mathcal A_i^{\rm scalar}(X_i)
    =
    \sum_{j=0}^{K-1} w_{i,j}u_{i-j},
    \qquad
    [\mathcal A_i^{\rm scalar}(X_i)]_p
    =
    \sum_{j=0}^{K-1} w_{i,j} I_C\,u_{i-j}[p],
    \label{eq:op_scalar}
\end{equation}
where \(I_C\) is the channel-wise identity matrix.
A more expressive choice is a pointwise operator, which learns channel mixing over the buffered velocities at each spatial position:
\begin{equation}
    [\mathcal A_i^{\rm pt}(X_i)]_p
    =
    a_i^{\rm pt}\bigl(X_i[p]\bigr),
    \qquad
    X_i[p]=(u_i[p],u_{i-1}[p],\ldots,u_{i-K+1}[p]).
    \label{eq:op_pointwise}
\end{equation}
A \(1{\times}1\) convolution is an instance of this class, which can mix channels and velocity history, so its output need not lie in the global span \(\mathcal V_i\). 
However, it remains pointwise, $\ie$, the update at \(p\) cannot depend on the buffer values at any \(q\neq p\).
We finally consider a spatial operator, where the update at \(p\) can depend on a neighborhood of positions:
\begin{equation}
    [\mathcal A_i^{\rm sp}(X_i)]_p
    =
    a_i^{\rm sp}\bigl(\{X_i[q]\}_{q\in\mathcal N(p)}\bigr),
    \label{eq:op_spatial}
\end{equation}
with \(\mathcal N(p)\) denoting the spatial support of the operator.
Convolutional kernels instantiate this class by enabling local \(q\to p\) interactions while keeping the operator lightweight.

\vspace{-6pt}
\paragraph{Nested expressiveness.}
These update classes form a natural hierarchy.
Scalar-coefficient solvers are a special case of pointwise operators, and pointwise operators are a special case of spatial operators with \(\mathcal N(p)=\{p\}\).
Thus, if \(\mathcal S_{\rm scalar}\), \(\mathcal S_{\rm pt}\), and \(\mathcal S_{\rm sp}\) denote the corresponding solver families,
\begin{equation}
    \mathcal S_{\rm scalar}
    \subseteq
    \mathcal S_{\rm pt}
    \subseteq
    \mathcal S_{\rm sp}.
    \label{eq:solver_family_nesting}
\end{equation}
The best achievable teacher-matching objective is non-increasing as the update family enlarges:
\begin{equation}
    \inf_{S\in\mathcal S_{\rm sp}} \mathcal J(S)
    \;\leq\;
    \inf_{S\in\mathcal S_{\rm pt}} \mathcal J(S)
    \;\leq\;
    \inf_{S\in\mathcal S_{\rm scalar}} \mathcal J(S).
    \label{eq:hierarchy_bound}
\end{equation}

\paragraph{Why spatial operators?}
\begin{wraptable}{r}{0.42\linewidth}
\centering
\caption{\textbf{Operator ablation}. SANA-0.6B, NFE=5, evaluated on MJHQ-25K.}
\label{tab:spatial_mixing_ablation}
\setlength{\tabcolsep}{5pt}
\resizebox{\linewidth}{!}{%
\begin{tabular}{@{}ll|cc@{}}
\toprule
 Operator & Kernel & \textbf{FID$\downarrow$} & \textbf{ImageReward$\uparrow$} \\
\midrule
Scalar & DyWeight & 9.05 & 0.8856 \\
Pointwise & $1{\times}1$            & 8.97 & 0.8878 \\
\midrule
\multirow{3}{*}{Spatial}
 & $3{\times}3$            & 8.22 & 0.9462 \\
 & $5{\times}5$            & 8.21 & 0.9226 \\
 & $9{\times}9$            & 8.16 & 0.9285 \\
\bottomrule
\end{tabular}%
}
\vspace{-10pt}
\end{wraptable}

The operator hierarchy above gives several ways to enlarge the scalar-coefficient family.
We examine the local ODE structure behind the out-of-span residual to guide the choice of operator class.
For \(\dot x(t)=u(x(t),t)\), the derivative of the velocity field along the trajectory is $D_tu =\partial_tu + J_xu\,u,
\label{eq:main_material_derivative}$
and for a spatial signal, the Jacobian--velocity product has the form
$[J_xu\,u]_p= \sum_q \frac{\partial u[p]}{\partial x[q]}u[q]. \label{eq:main_jvp_spatial}$
When off-diagonal sensitivities \(\partial u[p]/\partial x[q]\) are non-negligible, the correction at \(p\) depends on information from other positions.
Projection onto \(\mathcal V_i^\perp\) removes only components explained by the velocity span, so spatially coupled components that are not already in \(\mathcal V_i\) can remain in the out-of-span residual.
Since a spatial operator can directly access neighboring buffer entries, we use spatial kernels as a lightweight inductive bias for modeling the out-of-span residual.

Table~\ref{tab:spatial_mixing_ablation} compares scalar-coefficient updates, pointwise \(1{\times}1\) operators, and spatial convolutional operators under the same training protocol.
Using spatial kernels consistently outperform per-position alternatives, supporting our choice of local convolution as the operator instantiation.

\section{SpanLift: An operator-augmented neural solver}
\label{sec:operator_solver}

The previous section motivates enlarging scalar-coefficient solvers with an operator that can access directions outside the velocity-buffer span.
\textcolor{mod}{We instantiate the operator view above as \textbf{SpanLift}, an operator-augmented neural solver built on top of a strong scalar-coefficient base solver.}
The base solver provides a reliable in-span update, while the learned operator supplies additional directions beyond the velocity-buffer span.

\paragraph{Residual operator parameterization.}
Let \(X_i=(u_i,u_{i-1},\ldots,u_{i-K+1})\) denote the velocity buffer from Section~\ref{sec:operator_hierarchy}.
We write the base update as $\Delta_i^{\rm base} = \mathcal A_i^{\rm base}(X_i),$
where \(\mathcal A_i^{\rm base}\) is an analytic or learned scalar-coefficient solver, such as~\eqref{eq:op_scalar}.
SpanLift augments this update with a learned residual operator:
\begin{equation}
    \Delta_i^{\rm SL}
    =
    \alpha_i\,\Delta_i^{\rm base}
    +
    \beta_i\,\Delta_i^{\rm op},
    \qquad
    \Delta_i^{\rm op}
    =
    \mathcal R_\phi(\widetilde X_i;\xi_i),
    \label{eq:residual_form}
\end{equation}
where \(\alpha_i\) and \(\beta_i\) are step-wise learnable scalars, \(\mathcal R_\phi\) is a learned spatial residual operator, and
$\widetilde X_i := \operatorname{Concat}(x_i,u_i,u_{i-1},\ldots,u_{i-K+1})
\label{eq:operator_input}$
is the residual-operator input.
The conditioning variable \(\xi_i=[t_i,\lambda_i,h_i]\) provides the current time, logSNR, and step size.
We initialize the residual branch close to the base solver, $\eg$, with \(\alpha_i=1\) and \(\beta_i\) set to zero or a small value.

This parameterization separates the role of the base solver and the learned residual operator.
The base update provides a strong in-span direction inherited from existing multistep solvers.
The residual operator is not constrained to lie in \(\mathcal V_i\), and its spatial support enables the local \(q\to p\) interactions in~\eqref{eq:op_spatial}.
We do not enforce \(\Delta_i^{\rm op}\perp\mathcal V_i\), nor do we supervise it with the local residual \(r_i^\perp\).
Instead, the endpoint objective determines how the operator uses its extra degrees of freedom across the rollout.
The base solver acts as a solver prior, so the learned operator only needs to model the remaining endpoint error while retaining the inductive bias of existing multistep solvers.

At each step, we evaluate the frozen generative model once to obtain \(u_i\), form \(\Delta_i^{\rm base}\) from the velocity buffer, compute \(\Delta_i^{\rm op}\) using \(\mathcal R_\phi\), and update
\begin{equation}
    x_{i+1}
    =
    x_i
    +
    h_i
    \bigl(
        \alpha_i\Delta_i^{\rm base}
        +
        \beta_i\Delta_i^{\rm op}
    \bigr).
    \label{eq:ours_update}
\end{equation}
Since \(\mathcal R_\phi\) does not call the pretrained velocity model, the method does not increase model NFE; it only adds a lightweight neural overhead.

\paragraph{Architecture.}
We instantiate \(\mathcal R_\phi\) as a shallow FiLM-conditioned CNN.
A \(1{\times}1\) stem first mixes the channel dimension of \(\widetilde X_i\).
Then, FiLM-modulated residual blocks with depthwise spatial convolutions provide the spatial mixing motivated in Section~\ref{sec:operator_hierarchy}.
A final convolutional head maps the features to the residual update \(\Delta_i^{\rm op}\).
The default configuration uses hidden dimension \(d=64\) with two residual blocks, yielding negligible parameter overhead relative to the frozen diffusion or flow model.
Full architectural details are provided in Appendix A.

\paragraph{Training objective.}
We train SpanLift end-to-end through the full \(N\)-step rollout using the teacher-matching objective \(\mathcal J\) in~\eqref{eq:global_objective}.
Thus, gradients reflect the cumulative effect of each residual update on the final output, rather than supervising each step with a fixed local target.
Following prior learned-solver methods~\cite{LD3,s4s,dyweight}, we choose the per-sample loss \(\ell\) according to the model type.
\vspace{-2pt}

For pixel-space diffusion models~\cite{EDM}, we use an Inception loss~\cite{inception}.
For latent flow-matching models~\cite{flux,xie2025sana,gao2023prediff}, the increased capacity of the spatial residual operator makes loss design more important.
Because scalar solvers have limited capacity, they are less prone to fitting fine-grained latent \(\ell_2\) discrepancies.
A higher-capacity operator can match these details more directly, which may reduce latent MSE while degrading perceptual sharpness.
We therefore supplement latent matching with an image-space perceptual term.
Specifically, let \(z_1^\star\) and \(\hat z_1^\phi\) denote the final latents from teacher and SpanLift, and let \(D\) denote the frozen decoder used only to compute image-space perceptual features.
We use
\begin{equation}
    \ell(\hat z_1^\phi,z_1^\star)
    =
    c\,\|\hat z_1^\phi-z_1^\star\|^2
    +
    \bigl\|
        f_{\rm DINO}(D(\hat z_1^\phi))
        -
        f_{\rm DINO}(D(z_1^\star))
    \bigr\|^2,
    \label{eq:dino_loss}
\end{equation}
where \(f_{\rm DINO}\) denotes a frozen DINOv2~\cite{oquab2023dinov2} and \(c=0.1\).
Empirically, this perceptual term improves visual sharpness relative to latent MSE alone.
Note that we use the identical loss~\eqref{eq:dino_loss} for DyWeight.

\makeatletter
\renewcommand\thesubtable{(\alph{subtable})}
\makeatother

\begin{table*}[t]
\caption{
\textbf{Image generation with pixel space diffusion model}. FID$\downarrow$ \cite{fid} results across four datasets, including {(a)} CIFAR-10~\cite{cifar}, 
{(b)} ImageNet~\cite{imagenet},
{(c)} FFHQ~\cite{ffhq}, and
{(d)} AFHQv2~\cite{afhqv2}.
We compare SpanLift with handcrafted multi-step solvers and learnable solvers.
}
\label{tab:main_results}
\vspace{-1mm}

{\small
\setlength{\tabcolsep}{3.2pt}
\renewcommand{\arraystretch}{1.08}
\captionsetup[subfloat]{labelformat=simple, labelsep=space}

\begin{minipage}[t]{0.485\textwidth}
\centering
\fontsize{7.5}{9}\selectfont

\subfloat[Unconditional gen. on \textbf{CIFAR-10} $32\!\times\!32$.]{
\begin{tabular}{@{}cp{2.55cm}cccc@{}}
\toprule
& \multirow{2}{*}{Method} & \multicolumn{4}{c}{NFE} \\
\cmidrule(l){3-6}
& & 3 & 4 & 5 & 6 \\
\midrule
\multirow{3}{*}{\rotatebox{90}{\fontsize{6.0pt}{7.3pt}\selectfont Handcrafted}}
& iPNDM(3M) \cite{deis}           & 24.55 & 13.92 & 7.77 & 5.07 \\
& DPMPP(3M)~\cite{dpmpp} & 55.76 & 22.41 & 9.94 & 5.97 \\
& UniPC(3M)~\cite{UniPC}            & 109.60 & 45.20 & 23.98 & 11.14 \\
\midrule
\multirow{9}{*}{\rotatebox{90}{\fontsize{6.3pt}{7.3pt}\selectfont Learnable}}
& AMED-Solver~\cite{AMED}      & 18.49 & 17.18 & 7.59 & 7.04 \\
& AMED-Plugin~\cite{AMED}      & 10.81 & - & 6.61 & - \\
& EPD-Solver~\cite{EPD}       & 10.40 & - & 4.33 & - \\
& EPD-Plugin~\cite{EPD}      & 10.54 & - & 4.47 & - \\
& LD3~\cite{LD3}              & 16.52 & 9.31 & 6.39 & 3.35 \\
& DLMS~\cite{DLMS}             & - & 4.52 & 3.23 & 2.81 \\
& S4S-Alt~\cite{s4s}          & 16.95 & 6.35 & 3.73 & 2.67 \\
& DyWeight~\cite{dyweight}
                   & 8.16 & 3.92 & 3.02 & 2.61 \\
& \cellcolor{sotacolor}SpanLift
                   & \cellcolor{sotacolor}\textbf{5.69}
                   & \cellcolor{sotacolor}\textbf{3.57}
                   & \cellcolor{sotacolor}\textbf{2.76}
                   & \cellcolor{sotacolor}\textbf{2.41} \\
\bottomrule
\end{tabular}
}

\vspace{0.6em}

\subfloat[Conditional gen. on \textbf{ImageNet} $64\!\times\!64$.]{
\begin{tabular}{@{}cp{2.55cm}cccc@{}}
\toprule
& \multirow{2}{*}{Method} & \multicolumn{4}{c}{NFE} \\
\cmidrule(l){3-6}
& & 3 & 4 & 5 & 6 \\
\midrule
\multirow{3}{*}{\rotatebox{90}{\fontsize{6.0pt}{7.3pt}\selectfont Handcrafted}}
& iPNDM(3M) \cite{deis}           & 34.81 & 21.33 & 15.54 & 10.27 \\
& DPMPP(3M)~\cite{dpmpp} & 65.19 & 30.56 & 16.87 & 11.38 \\
& UniPC(3M)~\cite{UniPC}            & 91.38 & 55.63 & 24.36 & 14.30 \\
\midrule
\multirow{8}{*}{\rotatebox{90}{\fontsize{6.0pt}{7.3pt}\selectfont Learnable}}
& AMED-Solver~\cite{AMED}      & 38.10 & 32.69 & 10.74 & 10.63 \\
& AMED-Plugin~\cite{AMED}      & 28.06 & - & 13.83 & - \\
& EPD-Solver~\cite{EPD}       & 18.28 & - & 6.35 & - \\
& EPD-Plugin~\cite{EPD}       & 19.89 & - & 8.17 & - \\
& LD3~\cite{LD3}              & 27.82 & 17.03 & 11.55 & 7.53 \\
& DLMS~\cite{DLMS}            & - & 10.07 & 7.16 & 7.08 \\
& DyWeight~\cite{dyweight}
                   & 17.37 & 9.62 & 6.30 & 6.15 \\
& \cellcolor{sotacolor}SpanLift
                   & \cellcolor{sotacolor}\textbf{11.83}
                   & \cellcolor{sotacolor}\textbf{6.42}
                   & \cellcolor{sotacolor}\textbf{4.71}
                   & \cellcolor{sotacolor}\textbf{3.78} \\
\bottomrule
\end{tabular}
}

\end{minipage}
\hfill
\begin{minipage}[t]{0.485\textwidth}
\centering
\fontsize{7.5}{9}\selectfont

\subfloat[Unconditional gen. on \textbf{FFHQ} $64\!\times\!64$.]{
\begin{tabular}{@{}cp{2.55cm}cccc@{}}
\toprule
& \multirow{2}{*}{Method} & \multicolumn{4}{c}{NFE} \\
\cmidrule(l){3-6}
& & 3 & 4 & 5 & 6 \\
\midrule
\multirow{3}{*}{\rotatebox{90}{\fontsize{6.0pt}{7.3pt}\selectfont Handcrafted}}
& iPNDM(3M) \cite{deis}           & 27.72 & 20.07 & 13.80 & 8.61 \\
& DPMPP(3M)~\cite{dpmpp} & 66.07 & 30.06 & 13.47 & 8.25 \\
& UniPC(3M)~\cite{UniPC}            & 86.43 & 44.78 & 21.40 & 12.85 \\
\midrule
\multirow{9}{*}{\rotatebox{90}{\fontsize{6.0pt}{7.3pt}\selectfont Learnable}}
& AMED-Solver~\cite{AMED}      & 47.31 & 26.89 & 14.80 & 9.97 \\
& AMED-Plugin~\cite{AMED}      & 26.87 & - & 12.49 & - \\
& EPD-Solver~\cite{EPD}       & 21.74 & - & 7.84 & - \\
& EPD-Plugin~\cite{EPD}       & 19.02 & - & 7.97 & - \\
& LD3~\cite{LD3}              & 23.86 & 17.96 & 10.36 & 5.97 \\
& DLMS~\cite{DLMS}             & - & 9.63 & 6.85 & 5.82 \\
& S4S-Alt~\cite{s4s}          & 19.86 & 10.63 & 6.25 & 4.62 \\
& DyWeight~\cite{dyweight}
                   & 16.78 & 9.17 & 5.85 & 3.93 \\
& \cellcolor{sotacolor}SpanLift
                   & \cellcolor{sotacolor}\textbf{10.18}
                   & \cellcolor{sotacolor}\textbf{5.77}
                   & \cellcolor{sotacolor}\textbf{3.87}
                   & \cellcolor{sotacolor}\textbf{3.11} \\
\bottomrule
\end{tabular}
}

\vspace{0.6em}

\subfloat[Unconditional gen. on \textbf{AFHQv2} $64\!\times\!64$ .]{
\begin{tabular}{@{}cp{2.55cm}cccc@{}}
\toprule
& \multirow{2}{*}{Method} & \multicolumn{4}{c}{NFE} \\
\cmidrule(l){3-6}
& & 3 & 4 & 5 & 6 \\
\midrule
\multirow{3}{*}{\rotatebox{90}{\fontsize{6.0pt}{7.3pt}\selectfont Handcrafted}}
& iPNDM(3M) \cite{deis}          & 15.53 & 8.73 & 5.58 & 3.81 \\
& DPMPP(3M)~\cite{dpmpp} & 35.05 & 21.04 & 10.63 & 6.24 \\
& UniPC(3M)~\cite{UniPC}           & 60.89 & 33.78 & 13.01 & 8.27 \\
\midrule
\multirow{4}{*}{\rotatebox{90}{\fontsize{6.3pt}{7.3pt}\selectfont Learnable}}
& AMED-Solver~\cite{AMED}      & 31.82 & 18.99 & 7.34 & 8.19 \\
& LD3~\cite{LD3}              & 17.94 & 9.96 & 6.09 & 3.63 \\
& S4S-Alt~\cite{s4s}          & 14.71 & 6.52 & 3.89 & 2.70 \\
& DyWeight~\cite{dyweight}    & 9.16  & 5.38 & 3.20 & 2.66 \\
& \cellcolor{sotacolor}SpanLift
                   & \cellcolor{sotacolor}\textbf{5.89}
                   & \cellcolor{sotacolor}\textbf{3.85}
                   & \cellcolor{sotacolor}\textbf{2.74}
                   & \cellcolor{sotacolor}\textbf{2.42} \\
\bottomrule
\end{tabular}
}

\end{minipage}
}
\vspace{-14pt}
\end{table*}
\begin{table}[t]
\begin{minipage}[t]{0.4\linewidth}
    \centering
    \caption{\textbf{Text-to-image generation on SANA-0.6B~\cite{xie2025sana}.} Evaluated on MJHQ-25K, CFG=4.5.}
    \label{tab:sana}
    \vspace{0pt}
    
    \fontsize{8}{10}\selectfont
    \setlength{\tabcolsep}{3pt}
    \resizebox{\linewidth}{!}{
    \begin{tabular}{clccc}
    \toprule
    Metrics & Method & NFE=3 & NFE=5 & NFE=7 \\
    \midrule
    \multirow{3}{*}{FID$\downarrow$} 
    & iPNDM(2M)~\cite{deis} & 75.92 & 20.12  & 10.21   \\
    & DyWeight~\cite{dyweight} & 19.82 & 9.05 & 7.54 \\
    & \cellcolor{sotacolor} SpanLift 
        & \cellcolor{sotacolor}\textbf{14.40} 
        & \cellcolor{sotacolor}\textbf{8.22} 
        & \cellcolor{sotacolor}\textbf{7.51} \\
    \midrule
    \multirow{3}{*}{ImageReward$\uparrow$} 
    & iPNDM(2M)~\cite{deis} & -1.306 & 0.2654 & 0.8368 \\
    & DyWeight~\cite{dyweight} & 0.1618 & 0.8856 & 1.020 \\
    & \cellcolor{sotacolor} SpanLift 
        & \cellcolor{sotacolor}\textbf{0.4350} 
        & \cellcolor{sotacolor}\textbf{0.9462} 
        & \cellcolor{sotacolor}\textbf{1.023} \\
    \bottomrule
    \end{tabular}
    }
\end{minipage}
\hfill
\begin{minipage}[t]{0.59\linewidth}
    \centering
    \captionof{table}{\textbf{Precipitation nowcasting on PreDiff~\cite{gao2023prediff}.} Report HSS, LPIPS, CSI on SEVIR\cite{sevir}.}
    \label{tab:prediff_samplers}
    \vspace{0pt}
    \resizebox{\linewidth}{!}{%
    \begin{tabular}{cl ccccc}
      \toprule
      Steps & Method & HSS \small $\uparrow$ & LPIPS \small $\downarrow$ & CSI \small $\uparrow$ & \small CSI-4 $\uparrow$ & \small CSI-16 $\uparrow$ \\
      \midrule
      \multicolumn{7}{l}{\textit{Reference}} \\
      1000 & DDPM~\cite{DDPM}                    & 0.5422 & 0.3210 & 0.3972 & 0.4495 & 0.6133 \\
       100 & DDIM~\cite{DDIM} (teacher)           & 0.5373 & 0.3199 & 0.3822 & 0.4419 & 0.6034 \\
      \midrule
      \multicolumn{7}{l}{\textit{Few-step (5 steps)}} \\
         5 & DDIM                             & 0.5376 & 0.3429 & 0.3897 & 0.4261 & 0.5137 \\
         \cellcolor{sotacolor}5 &\cellcolor{sotacolor}SpanLift (MSE, teacher)           &\cellcolor{sotacolor}0.5255     &\cellcolor{sotacolor}0.3261            &\cellcolor{sotacolor}0.3857  &\cellcolor{sotacolor}0.4431 &\cellcolor{sotacolor}0.5992  \\
         \cellcolor{sotacolor}5 &\cellcolor{sotacolor}SpanLift (Inc., teacher)    &\cellcolor{sotacolor}0.4660 &\cellcolor{sotacolor}0.3239          &\cellcolor{sotacolor}0.4050 &\cellcolor{sotacolor}0.4560 &\cellcolor{sotacolor}0.6079 \\
         \cellcolor{sotacolor}5 &\cellcolor{sotacolor}SpanLift (Inc., target)         &\cellcolor{sotacolor}\textbf{0.6009} &\cellcolor{sotacolor}\textbf{0.3065} &\cellcolor{sotacolor}\textbf{0.4211} &\cellcolor{sotacolor}\textbf{0.4797} &\cellcolor{sotacolor}\textbf{0.6312} \\
      \bottomrule
    \end{tabular}
    }%
\end{minipage}
\end{table}

\vspace{-7pt}
\section{Experiments}
\vspace{-7pt}
We evaluate SpanLift in diverse few-step sampling settings, covering image generation with pixel-space diffusion and latent flow-matching models, and precipitation nowcasting. 
Our experiments address three questions: \textbf{Q1.} Does SpanLift improve few-step sampling across models/domains? \textbf{Q2.} Can SpanLift serve as a general residual operator across base solvers? \textbf{Q3.} Does the learned correction behave as the intended span-lifting residual?
Regarding implementation details and more results, see the appendix.
\subsection{Few-step image generation}
\label{sec:exp1}
\paragraph{Experimental setup.}
For pixel-space generation, we use the official EDM~\cite{EDM} checkpoints on CIFAR-10~\cite{cifar}, ImageNet~\cite{imagenet}, FFHQ~\cite{ffhq}, and AFHQv2~\cite{afhqv2}. 
For latent flow matching, we evaluate SANA-0.6B~\cite{xie2025sana} at $512{\times}512$ resolution, with additional FLUX.1-dev~\cite{flux} results in the appendix. 
SpanLift trains only the lightweight residual operator $R_\phi$, while keeping the generative backbone and the base solver frozen.

For pixel-space experiments, we follow the DyWeight~\cite{dyweight} protocol, including the teacher solver and training set size, and report FID~\cite{fid} over 50K generated samples. 
For SANA, we split MJHQ-30K into 5K training and 25K evaluation prompts. 
SpanLift and DyWeight are trained on the same precomputed teacher trajectories using 20-NFE iPNDM(2M), CFG$=4.5$, and PAG disabled. 
We report FID and ImageReward~\cite{imagereward} on the 25K evaluation set. Full hyperparameters and architectural details are provided in the appendix.
\paragraph{Results.}
Tables~\ref{tab:main_results} and~\ref{tab:sana} summarize the quantitative results. 
SpanLift achieves the lowest FID across all pixel space benchmarks and on SANA-0.6B, with larger gains at lower NFE.
This pattern is consistent with our span perspective, where large steps can expose residual directions that are not well captured by scalar recombination, while the spatial residual operator can partially compensate for them. 
Figure~\ref{fig:sana_nfe5_qual} shows the qualitative results, where both analytic and learned scalar solvers fail to form recognizable scene structure, whereas SpanLift recovers coherent objects and layouts. 
Qualitative results on FLUX.1-dev further show improvements in local prompt details, including legible text on a cake topper and a correctly depicted hand-held phone (Figure~\ref{fig:flux_nfe9_qual}).
\vspace{-3pt}
\begin{figure}[t]
  \begin{minipage}[t]{0.48\linewidth}
    \vspace{0pt}
    \centering
    \captionof{table}{\textbf{Base solver analysis.} FID on ImageNet with and without the SpanLift across different base solvers.}
    \label{tab:edm_base_solver}
    \fontsize{7.3}{8.8}\selectfont
    \setlength{\tabcolsep}{4pt}
    \resizebox{\linewidth}{!}{%
    \begin{tabular}{@{}l l ccc@{}}
      \toprule
      Base solver & Method & NFE=7 & NFE=9 & NFE=11 \\
      \midrule
      \multirow{2}{*}{RK-Euler}
      & Base  & 21.28  & 15.27  &11.71  \\
      &  \cellcolor{sotacolor}+SpanLift & \cellcolor{sotacolor}\textbf{4.47}   & \cellcolor{sotacolor}\textbf{4.12} & \cellcolor{sotacolor}\textbf{3.58} \\
      \midrule
      \multirow{2}{*}{RK-Heun\textsuperscript{$\diamond$}}
      & Base  & 4.37 & 3.37 & 2.96\\
      &  \cellcolor{sotacolor}+SpanLift & \cellcolor{sotacolor}\textbf{3.02}  &\cellcolor{sotacolor}\textbf{2.81}   &\cellcolor{sotacolor}\textbf{2.80}   \\
      \midrule
      \multirow{2}{*}{iPNDM(3M) \cite{deis}}
      & Base  & 6.57 &4.61  &3.74  \\
      &  \cellcolor{sotacolor}+SpanLift & \cellcolor{sotacolor}\textbf{4.17} & \cellcolor{sotacolor}\textbf{3.68} & \cellcolor{sotacolor}\textbf{3.23} \\
      \midrule
      \multirow{2}{*}{DyWeight~\cite{dyweight}}
      & Base  & 3.52 & 2.95 & 2.76  \\
      &  \cellcolor{sotacolor}+SpanLift & \cellcolor{sotacolor}\textbf{3.29} & \cellcolor{sotacolor}\textbf{2.83} & \cellcolor{sotacolor}\textbf{2.70} \\
      \bottomrule
      \multicolumn{5}{l}{\textsuperscript{$\diamond$}\footnotesize Matched by update steps; actual NFE is 15/19/23.} \\
    \end{tabular}%
    }
  \end{minipage}
  \hfill
  \begin{minipage}[t]{0.5\linewidth}
    \vspace{0pt}
    \centering
    \includegraphics[width=\linewidth]{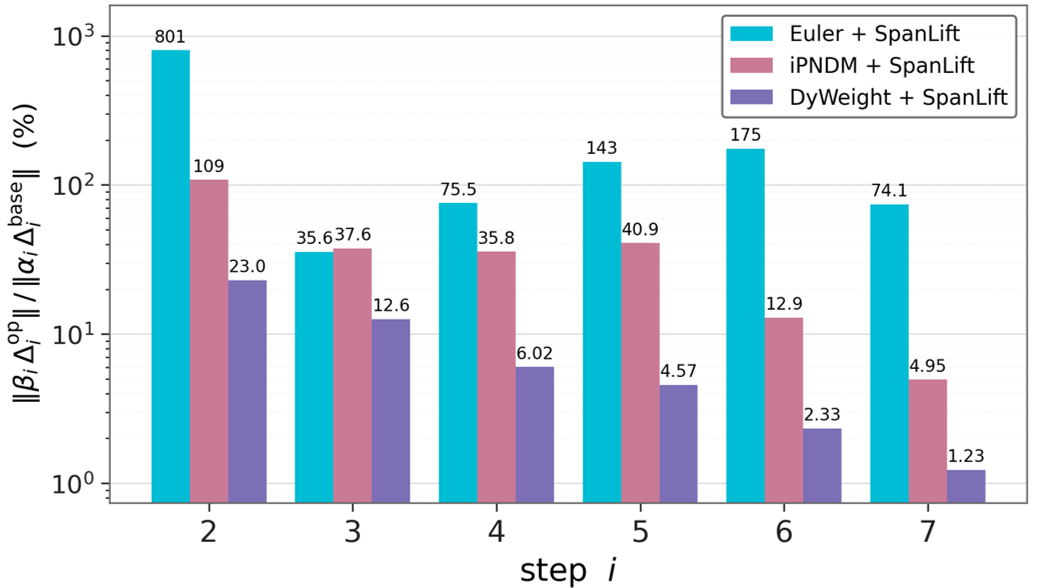}
    \caption{\textbf{Relative magnitude of the residual correction.} Step-wise $\|\beta_i\Delta_i^{\mathrm{op}}\| / \|\alpha_i\Delta_i^{\mathrm{base}}\|$. SpanLift applies proportionally larger corrections atop weaker base solvers.}
    \label{fig:ratio}
  \end{minipage}
\end{figure}
\vspace{-6pt}

\subsection{Precipitation nowcasting}
\label{sec:exp2}
\vspace{-3pt}
Experiment details are provided in Appendix A.
\vspace{-6pt}
\begin{figure}
    \centering
    \includegraphics[width=0.92\linewidth]{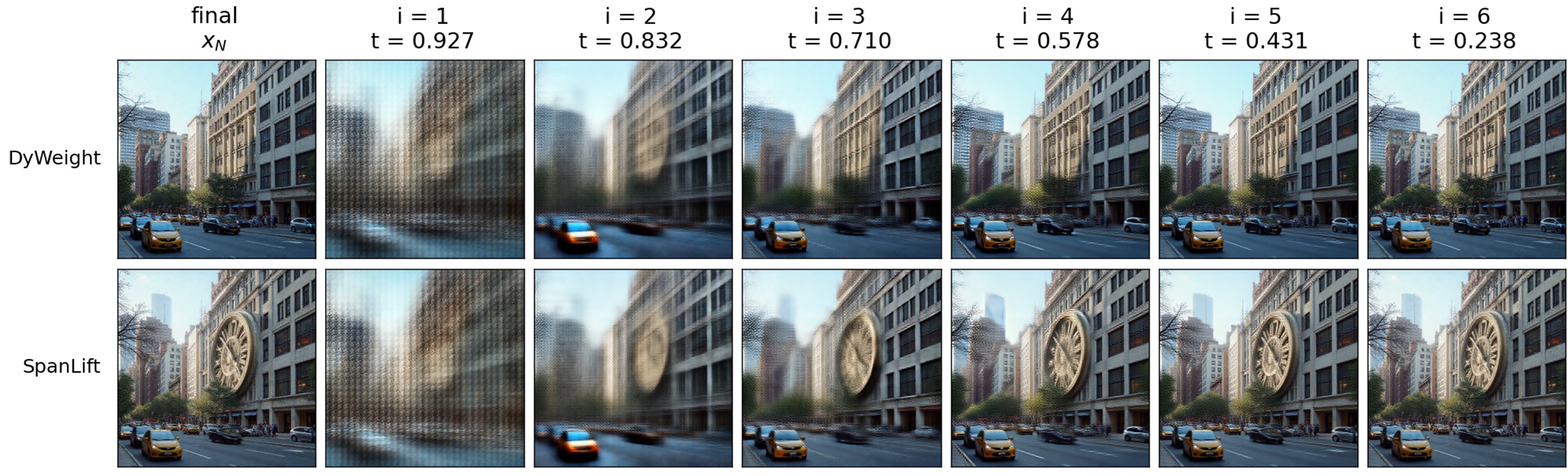}
    \caption{\textbf{Comparison on step-wise $\hat{x}_1$ estimation.}
    Starting from the same seed, DyWeight misses the key object specified in the prompt, whereas SpanLift corrects the later trajectory and produces a clear clock face.
    \textit{Prompt: ``A large clock on the side of a building above cars on the street.''}}
    \label{fig:correct_compare}
    \vspace{-11pt}
\end{figure}
\paragraph{Results.}
Table~\ref{tab:prediff_samplers} shows that reducing DDIM from 100 to 5 steps mainly damages perceptual quality and high-intensity precipitation structure, with LPIPS increasing from 0.3199 to 0.3429 and CSI-16 dropping from 0.6034 to 0.5137.
SpanLift recovers this lost structure at the same 5-step budget. With teacher supervision, the Inception-based loss improves CSI-16 to 0.6079, slightly surpassing the 100-step DDIM teacher.
With direct target supervision, SpanLift achieves the best score on every metric, including CSI-16 of 0.6312, exceeding even the 1000-step DDPM reference.

\vspace{-7.5pt}
\subsection{Analyses}
\label{sec:exp3}
\begin{figure*}[t]
  \begin{minipage}[t]{0.5\textwidth}
    \centering\small
    \captionof{table}{\textbf{Out-of-span ratio of $\Delta_i^{\mathrm{op}}$.} 
      Step-wise $\lVert P_i^{\perp}(\Delta_i^{\mathrm{op}})\rVert / 
      \lVert \Delta_i^{\mathrm{op}}\rVert$, where $P_i^\perp\!=\!I-\Pi_{\mathcal V_i}$.
      Higher values indicate more orthogonal to~$\mathcal{V}_i$.}
    \label{tab:offspan}
    {\footnotesize
    \begin{tabular}{c ccc}
      \toprule
      & \multicolumn{3}{c}{+SpanLift} \\
      \cmidrule(lr){2-4}
      Step $i$ & Euler & iPNDM & DyWeight \\
      \midrule
      2 & 0.97 & 0.92 & 0.90 \\
      3 & 0.85 & 0.88 & 0.92 \\
      4 & 0.97 & 0.86 & 0.95 \\
      5 & 0.96 & 0.85 & 0.95 \\
      6 & 0.94 & 0.93 & \textbf{0.98} \\
      7 & 0.85 & 0.97 & \textbf{0.99} \\
      \bottomrule
    \end{tabular}
    }
    \vspace{10pt}
    \caption{\textbf{Computational overhead analysis.} 
      Measured on ImageNet/EDM, 1$\times$A6000, batch=64.}
    \label{tab:overhead}
    \resizebox{\linewidth}{!}{%
    \begin{tabular}{lcc}
      \toprule
      Metric & DyWeight & SpanLift \\
      \midrule
      \# Params $\downarrow$ (M)          & 295.899 & 296.010 (+0.04\%) \\
      FLOPs / image $\downarrow$ (GFLOPs) & 1973.96 & 1978.75 (+0.24\%) \\
      Latency / image $\downarrow$ (ms)   & 49.74   & 51.29 (+3.1\%) \\
      Throughput $\uparrow$ (img/s)       & 20.10   & 19.50 ($-$3.0\%) \\
      \bottomrule
    \end{tabular}%
    }
  \end{minipage}%
  \hfill
  \begin{minipage}[t]{0.46\textwidth}
    \vspace{0pt}
    \centering
    \includegraphics[width=\linewidth]{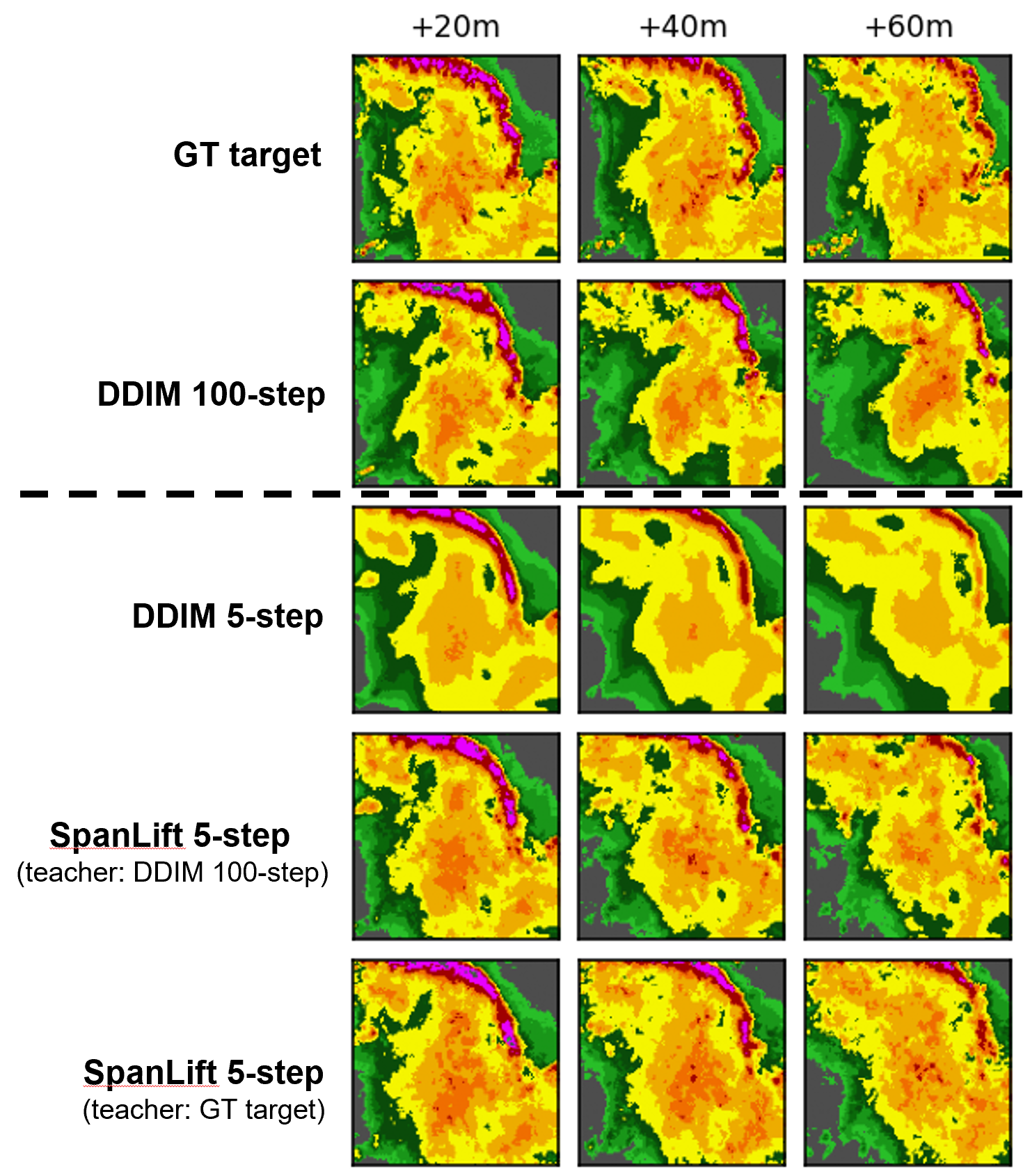}
    \caption{Qualitative results with PreDiff~\cite{gao2023prediff}.}
    \label{fig:prediff_qualitative}
  \end{minipage}
  \vspace{-18pt}
\end{figure*}
\vspace{-5.5pt}
All analyses in this section are conducted on ImageNet $64{\times}64$ using the EDM~\cite{EDM} checkpoint, and we provide additional analyses in Figure~\ref{fig:correct_compare} and Table~\ref{tab:overhead}.
\vspace{-7.5pt}
\paragraph{Base solver-agnostic correction.}
To answer Q2, we train SpanLift on four fixed base solvers
Table~\ref{tab:edm_base_solver} shows consistent FID improvements over Euler, Heun, iPNDM, and DyWeight.
The gains are largest for weaker bases such as Euler, but remain consistent even on the stronger DyWeight base.
Figure~\ref{fig:ratio} shows the relative magnitude of the residual correction, $\ie$, step-wise $\|\beta_i\Delta_i^{\mathrm{op}}\| / \|\alpha_i\Delta_i^{\mathrm{base}}\|$ after projecting the base and residual updates onto the subspace orthogonal to the teacher chord $x_1^{\mathrm{teacher}} - x_0$. 
Weaker bases require larger corrections, with Euler reaching $801\%$ of the projected base step and DyWeight staying below ${\sim}23\%$. 

\vspace{-7.5pt}
\paragraph{Out-of-span residual analysis.}
To answer Q3, we verify that the learned correction supplies directions outside the velocity-buffer span $\mathcal{V}_i$. Using $K{=}3$, we define $\mathcal{V}_i=\mathrm{span}\{u_i,u_{i-1},u_{i-2}\}$ and measure $\|P_i^\perp(\Delta_i^{\mathrm{op}})\|/\|\Delta_i^{\mathrm{op}}\|$, where $P_i^\perp=I-\Pi_{\mathcal{V}_i}$. 
As shown in Table~\ref{tab:offspan}, the ratio stays above $84\%$ for every step and base solver, indicating that $\Delta_i^{\mathrm{op}}$ is largely inaccessible to scalar recombination of the same velocity buffer. 
The ratios are highest for DyWeight, consistent with the view that a strong scalar base leaves mostly out-of-span residual structure for SpanLift to correct.

\vspace{-7.5pt}


\section{Conclusion}
\vspace{-8pt}
We introduced SpanLift, a lightweight operator-augmented neural solver that lifts scalar-coefficient updates beyond the velocity-buffer span. 
SpanLift keeps the backbone fixed and preserves the model NFE budget while expanding the solver update family. Empirical analyses show that the learned correction transfers across base solvers and supplies predominantly out-of-span directions. 
Across all evaluated settings, SpanLift achieves state-of-the-art performance under the extreme NFE budgets.

\bibliographystyle{unsrtnat}
\bibliography{references}

@String(IJCV = {Int. J. Comput. Vis.})

@String(CVPR= {IEEE Conf. Comput. Vis. Pattern Recog.})

@String(ICLR = {Int. Conf. Learn. Represent.})

@String(IJCV  = {IJCV})

@String(CVPR  = {CVPR})

@String(ICLR  = {ICLR})

@inproceedings{DDPM,
  title={Denoising diffusion probabilistic models},
  author={Ho, Jonathan and Jain, Ajay and Abbeel, Pieter},
  booktitle={NeurIPS},
  year={2020}
}

@inproceedings{DDIM,
  title={Denoising diffusion implicit models},
  author={Song, Jiaming and Meng, Chenlin and Ermon, Stefano},
  booktitle={ICLR},
  year={2021}
}

@inproceedings{UniPC,
  title={Unipc: A unified predictor-corrector framework for fast sampling of diffusion models},
  author={Zhao, Wenliang and Bai, Lujia and Rao, Yongming and Zhou, Jie and Lu, Jiwen},
  booktitle={NeurIPS},
  year={2023}
}

@article{EDM,
  title={Elucidating the design space of diffusion-based generative models},
  author={Karras, Tero and Aittala, Miika and Aila, Timo and Laine, Samuli},
  journal={NeurIPS},
  year={2022}
}

@inproceedings{PNDM,
  title={Pseudo Numerical Methods for Diffusion Models on Manifolds},
  author={Liu, Luping and Ren, Yi and Lin, Zhijie and Zhao, Zhou},
  booktitle={ICLR},
  year={2022}
}

@inproceedings{deis,
  title={Fast Sampling of Diffusion Models with Exponential Integrator},
  author={Zhang, Qinsheng and Chen, Yongxin},
  booktitle={ICLR},
  year={2023}
}

@inproceedings{tan2026stork,
  title={STORK: Faster Diffusion and Flow Matching Sampling by Resolving both Stiffness and Structure-Dependence},
  author={Tan, Zheng and Wang, Weizhen and Bertozzi, Andrea L. and Ryu, Ernest K.},
  booktitle={ICLR},
  year={2026}
}

@inproceedings{dpmsolver,
  title={Dpm-solver: A fast ode solver for diffusion probabilistic model sampling in around 10 steps},
  author={Lu, Cheng and Zhou, Yuhao and Bao, Fan and Chen, Jianfei and Li, Chongxuan and Zhu, Jun},
  booktitle={NeurIPS},
  year={2022}
}

@article{dpmpp,
  title={Dpm-solver++: Fast solver for guided sampling of diffusion probabilistic models},
  author={Lu, Cheng and Zhou, Yuhao and Bao, Fan and Chen, Jianfei and Li, Chongxuan and Zhu, Jun},
  journal={Machine Intelligence Research},
  year={2025},
  publisher={Springer}
}

@inproceedings{AMED,
  title={Fast ode-based sampling for diffusion models in around 5 steps},
  author={Zhou, Zhenyu and Chen, Defang and Wang, Can and Chen, Chun},
  booktitle={CVPR},
  year={2024}
}

@inproceedings{LD3,
  title={Learning to discretize denoising diffusion odes},
  author={Tong, Vinh and Trung-Dung, Hoang and Liu, Anji and Broeck, Guy Van den and Niepert, Mathias},
  booktitle={ICLR},
  year={2025}
}

@inproceedings{DLMS,
  title={Linear Multistep Solver Distillation for Fast Sampling of Diffusion Models},
  author={Liang, Yuchen and Fang, Xiangzhong and Chen, Hanting and Wang, Yunhe},
  booktitle={ICLR},
  year={2025}
}

@inproceedings{EPD,
  title={Distilling parallel gradients for fast ode solvers of diffusion models},
  author={Zhu, Beier and Wang, Ruoyu and Zhao, Tong and Zhang, Hanwang and Zhang, Chi},
  booktitle={CVPR},
  year={2025}
}

@inproceedings{DSS,
  title={Differentiable Solver Search for Fast Diffusion Sampling},
  author={Wang, Shuai and Li, Zexian and Song, Tianhui and Li, Xubin and Ge, Tiezheng and Zheng, Bo and Wang, Limin and others},
  booktitle={ICML},
  year={2025}
}

@article{cifar,
  title={Learning multiple layers of features from tiny images},
  author={Krizhevsky, Alex and Hinton, Geoffrey and others},
  year={2009},
  publisher={Toronto, ON, Canada}
}

@inproceedings{afhqv2,
  title={Stargan v2: Diverse image synthesis for multiple domains},
  author={Choi, Yunjey and Uh, Youngjung and Yoo, Jaejun and Ha, Jung-Woo},
  booktitle={CVPR},
  year={2020}
}

@inproceedings{ffhq,
  title={A style-based generator architecture for generative adversarial networks},
  author={Karras, Tero and Laine, Samuli and Aila, Timo},
  booktitle={CVPR},
  year={2019}
}

@inproceedings{imagenet,
  title={Imagenet large scale visual recognition challenge},
  author={Russakovsky, Olga and Deng, Jia and Su, Hao and Krause, Jonathan and Satheesh, Sanjeev and Ma, Sean and Huang, Zhiheng and Karpathy, Andrej and Khosla, Aditya and Bernstein, Michael and others},
  booktitle={IJCV},
  year={2015},
}

@inproceedings{gits,
  title={On the Trajectory Regularity of ODE-based Diffusion Sampling},
  author={Chen, D and Zhou, Z and Wang, C and Shen, C and Lyu, S},
  booktitle={ICML},
  year={2024},
}

@inproceedings{learningfast,
  title={Learning fast samplers for diffusion models by differentiating through sample quality},
  author={Watson, Daniel and Chan, William and Ho, Jonathan and Norouzi, Mohammad},
  booktitle={ICLR},
  year={2021}
}

@inproceedings{bns,
  title={Bespoke non-stationary solvers for fast sampling of diffusion and flow models},
  author={Shaul, Neta and Singer, Uriel and Chen, Ricky TQ and Le, Matthew and Thabet, Ali and Pumarola, Albert and Lipman, Yaron},
  booktitle={ICML},
  year={2024}
}

@inproceedings{shaul2024bespoke,
  title={Bespoke Solvers for Generative Flow Models},
  author={Shaul, Neta and P{\'e}rez, Juan C. and Chen, Ricky T. Q. and Thabet, Ali and Pumarola, Albert and Lipman, Yaron},
  booktitle={ICLR},
  year={2024}
}

@inproceedings{dyweight,
  title={DyWeight: Dynamic Gradient Weighting for Few-Step Diffusion Sampling},
  author={Tong Zhao and Mingkun Lei and Liangyu Yuan and Yanming Yang and Chenxi Song and Yang Wang and Beier Zhu and Chi Zhang},
  booktitle={CVPR},
  year={2026}
}

@misc{flux,
    author={Black Forest Labs},
    title={FLUX},
    year={2024},
    howpublished={\url{https://github.com/black-forest-labs/flux}},
}

@article{fid,
  title={Gans trained by a two time-scale update rule converge to a local nash equilibrium},
  author={Heusel, Martin and Ramsauer, Hubert and Unterthiner, Thomas and Nessler, Bernhard and Hochreiter, Sepp},
  journal={NeurIPS},
  year={2017}
}

@inproceedings{DMN,
  title={Accelerating diffusion sampling with optimized time steps},
  author={Xue, Shuchen and Liu, Zhaoqiang and Chen, Fei and Zhang, Shifeng and Hu, Tianyang and Xie, Enze and Li, Zhenguo},
  booktitle={CVPR},
  year={2024}
}

@inproceedings{timesteptuner,
  title={Towards more accurate diffusion model acceleration with a timestep tuner},
  author={Xia, Mengfei and Shen, Yujun and Lei, Changsong and Zhou, Yu and Zhao, Deli and Yi, Ran and Wang, Wenping and Liu, Yong-Jin},
  booktitle={CVPR},
  year={2024}
}

@inproceedings{s4s,
  title={S4S: Solving for a diffusion model solver},
  author={Frankel, Eric and Chen, Sitan and Li, Jerry and Koh, Pang Wei and Ratliff, Lillian J and Oh, Sewoong},
  booktitle={ICML},
  year={2025}
}

@inproceedings{ays,
  title={Align Your Steps: Optimizing Sampling Schedules in Diffusion Models},
  author={Sabour, Amirmojtaba and Fidler, Sanja and Kreis, Karsten},
  booktitle={ICML},
  year={2024}
}

@inproceedings{D-ODE,
  title={Distilling ODE Solvers of Diffusion Models into Smaller Steps},
  author={Kim, Sanghwan and Tang, Hao and Yu, Fisher},
  booktitle={CVPR},
  year={2024}
}

@inproceedings{inception,
  title={Going deeper with convolutions},
  author={Szegedy, Christian and Liu, Wei and Jia, Yangqing and Sermanet, Pierre and Reed, Scott and Anguelov, Dragomir and Erhan, Dumitru and Vanhoucke, Vincent and Rabinovich, Andrew},
  booktitle={CVPR},
  year={2015}
}

@article{imagereward,
  title={Imagereward: Learning and evaluating human preferences for text-to-image generation},
  author={Xu, Jiazheng and Liu, Xiao and Wu, Yuchen and Tong, Yuxuan and Li, Qinkai and Ding, Ming and Tang, Jie and Dong, Yuxiao},
  journal={NeurIPS},
  year={2023}
}

@inproceedings{park2026dualsolver,
  title={Dual-Solver: A Generalized {ODE} Solver for Diffusion Models with Dual Prediction},
  author={Park, Soochul and Lee, Yeon Ju},
  booktitle={ICLR},
  year={2026},
}

@inproceedings{min2026bezierflow,
  title={B{\'e}zierFlow: Learning B{\'e}zier Stochastic Interpolant Schedulers for Few-Step Generation},
  author={Min, Yunhong and Koo, Juil and Yoo, Seungwoo and Sung, Minhyuk},
  booktitle={ICLR},
  year={2026},
}

@misc{oquab2023dinov2,
  title={DINOv2: Learning Robust Visual Features without Supervision},
  author={Oquab, Maxime and Darcet, Timoth{\'e}e and Moutakanni, Theo and Vo, Huy and Szafraniec, Marc and Khalidov, Vasil and Fernandez, Pierre and Haziza, Daniel and Massa, Francisco and El-Nouby, Alaaeldin and others},
  journal={arXiv preprint arXiv:2304.07193},
  year={2023}
}

@inproceedings{gao2023prediff,
  title={PreDiff: Precipitation Nowcasting with Latent Diffusion Models},
  author={Gao, Zhihan and Shi, Xingjian and Han, Boran and Wang, Hongya and Jin, Xiaoyong and Maddix, Danielle C. and Zhu, Yi and Li, Mu and Wang, Yuyang},
  booktitle={NeurIPS},
  year={2023},
}

@inproceedings{xie2025sana,
  title={{SANA}: Efficient High-Resolution Image Synthesis with Linear Diffusion Transformers},
  author={Xie, Enze and Chen, Junsong and Chen, Junyu and Cai, Han and Tang, Haotian and Lin, Yujun and Zhang, Zhekai and Li, Muyang and Zhu, Ligeng and Lu, Yao and Han, Song},
  booktitle={ICLR},
  year={2025},
}

@book{hairer1993solving,
  title={Solving Ordinary Differential Equations I: Nonstiff Problems},
  author={Hairer, Ernst and N{\o}rsett, Syvert P and Wanner, Gerhard},
  edition={2nd},
  publisher={Springer-Verlag},
  address={Berlin Heidelberg},
  year={1993}
}

@article{watson2023rfdiffusion,
  title={De novo design of protein structure and function with {RFdiffusion}},
  author={Watson, Joseph L and Juergens, David and Bennett, Nathaniel R and Trippe, Brian L and Yim, Jason and Eisenach, Helen E and others},
  journal={Nature},
  volume={620},
  pages={1089--1100},
  year={2023}
}

@inproceedings{salimans2022pd,
  title={Progressive distillation for fast sampling of diffusion models},
  author={Salimans, Tim and Ho, Jonathan},
  booktitle={ICLR},
  year={2022}
}

@inproceedings{song2023cm,
  title={Consistency models},
  author={Song, Yang and Dhariwal, Pankaj and Chen, Mark and Sutskever, Ilya},
  booktitle={ICML},
  year={2023}
}

@inproceedings{song2024ict,
  title={Improved techniques for training consistency models},
  author={Song, Yang and Dhariwal, Pankaj},
  booktitle={ICLR},
  year={2024}
}

@article{wan2025,
  title={Wan: Open and advanced large-scale video generative models},
  author={Team Wan and Wang, Ang and Ai, Baole and others},
  journal={arXiv preprint arXiv:2503.20314},
  year={2025}
}

@inproceedings{bose2024foldflow,
  title={{SE}(3)-Stochastic Flow Matching for Protein Backbone Generation},
  author={Bose, Avishek Joey and Akhound-Sadegh, Tara and Fatras, Kilian and Huguet, Guillaume and Rector-Brooks, Jarrid and Liu, Cheng-Hao and Nica, Andrei Cristian and Rokkum, Maksym and Wolf, Guy and Tong, Alexander},
  booktitle={ICLR},
  year={2024},
}

@article{kharazmi2021hp,
  title={hp-VPINNs: Variational physics-informed neural networks with domain decomposition},
  author={Kharazmi, Ehsan and Zhang, Zhongqiang and Karniadakis, George Em},
  journal={Computer Methods in Applied Mechanics and Engineering},
  year={2021},
  publisher={Elsevier}
}

@inproceedings{palpinns,
  title={PINNs with Learnable Quadrature},
  author={Pal, Sourav and Azizzadenesheli, Kamyar and Singh, Vikas},
  booktitle={NeurIPS},
  year      = {2025},
}

@article{sevir,
  title={Sevir: A storm event imagery dataset for deep learning applications in radar and satellite meteorology},
  author={Veillette, Mark and Samsi, Siddharth and Mattioli, Chris},
  booktitle={NeurIPS},
  year={2020}
}

\end{document}